\documentclass[runningheads]{llncs}
\usepackage{graphicx}
\usepackage[dvipsnames]{xcolor} % additional package
\usepackage{tikz}
\usepackage{comment}
\usepackage{amsmath,amssymb} % define this before the line numbering.
\usepackage{color}

\usepackage[accsupp]{axessibility}  % Improves PDF readability for those with disabilities.

\usepackage{cite}
\usepackage[capitalize]{cleveref}
\usepackage[caption=false]{subfig}

\begin{document}
% \renewcommand\thelinenumber{\color[rgb]{0.2,0.5,0.8}\normalfont\sffamily\scriptsize\arabic{linenumber}\color[rgb]{0,0,0}}
% \renewcommand\makeLineNumber {\hss\thelinenumber\ \hspace{6mm} \rlap{\hskip\textwidth\ \hspace{6.5mm}\thelinenumber}}
% \linenumbers
\pagestyle{headings}
\mainmatter
\def\ECCVSubNumber{7288}  % Insert your submission number here

\title{Pure Transformer with Integrated Experts for Scene Text Recognition} % Replace with your title

% INITIAL SUBMISSION 
\begin{comment}
\titlerunning{ECCV-22 submission ID \ECCVSubNumber} 
\authorrunning{ECCV-22 submission ID \ECCVSubNumber} 
\author{Anonymous ECCV submission}
\institute{Paper ID \ECCVSubNumber}
\end{comment}
%******************

% CAMERA READY SUBMISSION
%\begin{comment}
\titlerunning{PTIE for STR}
% If the paper title is too long for the running head, you can set
% an abbreviated paper title here
%
% \author{Yew Lee Tan\inst{1}\orcidlink{0000-0002-8117-105X}\index{Tan, Yew Lee} \and
% Adams Wai-Kin Kong\inst{1}\orcidlink{0000-0002-9728-9511}\index{Kong, Adams Wai-Kin} \and
% Jung-Jae Kim\inst{2}\orcidlink{ 0000-0001-6024-0047}}
\author{Yew Lee Tan\inst{1} \and
Adams Wai-Kin Kong\inst{1} \and
Jung-Jae Kim\inst{2}}
\authorrunning{Y.L. Tan et al.}
% First names are abbreviated in the running head.
% If there are more than two authors, 'et al.' is used.
%
\institute{Nanyang Technological University, Singapore
%\email{lncs@springer.com}\\
\and
Institute for Infocomm Research, A*STAR, Singapore}
%\email{\{abc,lncs\}@uni-heidelberg.de}}
%\end{comment}
%******************
\maketitle

\begin{abstract}
Scene text recognition (STR) involves the task of reading text in cropped images of natural scenes. Conventional models in STR employ convolutional neural network (CNN) followed by recurrent neural network in an encoder-decoder framework. In recent times, the transformer architecture is being widely adopted in STR as it shows strong capability in capturing long-term dependency which appears to be prominent in scene text images. Many researchers utilized transformer as part of a hybrid CNN-transformer encoder, often followed by a transformer decoder. However, such methods only make use of the long-term dependency mid-way through the encoding process. Although the vision transformer (ViT) is able to capture such dependency at an early stage, its utilization remains largely unexploited in STR. This work proposes the use of a transformer-only model as a simple baseline which outperforms hybrid CNN-transformer models. Furthermore, two key areas for improvement were identified. Firstly, the first decoded character has the lowest prediction accuracy. Secondly, images of different original aspect ratios react differently to the patch resolutions while ViT only employ one fixed patch resolution. To explore these areas, Pure Transformer with Integrated Experts (PTIE) is proposed. PTIE is a transformer model that can process multiple patch resolutions and decode in both the original and reverse character orders. It is examined on 7 commonly used benchmarks and compared with over 20 state-of-the-art methods. The experimental results show that the proposed method outperforms them and obtains state-of-the-art results in most benchmarks.
\keywords{transformer, scene text recognition, integrated experts}
\end{abstract}

\section{Introduction}
\label{sec:introduction}
Scene text recognition (STR) is useful in a wide array of applications such as document retrieval \cite{tsai2011mobile}, robot navigation \cite{schulz2015robot}%, industrial automation \cite{chowdhury2013extracting}
, and product recognition \cite{long2021scene}. Furthermore, STR is able to improve the lives of visually impaired by providing them access to visual information through texts encountered in natural scenes \cite{gurari2018vizwiz, ezaki2005improved}.

Traditionally, convolutional neural network (CNN) was used as a backbone in the encoder-decoder framework of STR to extract and encode features from the images \cite{chen2020text}. Recurrent neural network (RNN) was then used to capture sequence dependency and decode the features into a sequence of characters. In recent times, transformer \cite{vaswani2017attention} has been employed in STR models because of its strong capability in capturing long-term dependency. Some researchers have designed transformer-inspired modules \cite{zhang2021scene, wang2021two}, while others have utilized it as a hybrid CNN-transformer encoder \cite{fang2021read} and/or a transformer decoder in STR \cite{lee2020recognizing,lu2021master}.
%while others have adopted it as a decoder \cite{lu2021master}, a hybrid CNN-transformer encoder \cite{fang2021read}, or both \cite{lee2020recognizing}. 

Scene text usually has the same font, color, and style, thus exhibiting a coherent pattern. These properties suggest that STR has strong long-term dependency. Henceforth, recent works based on hybrid CNN-transformer \cite{fang2021read} outperform models with traditional architectures like CNN and RNN. A natural following question to ask is –- will STR performance be improved by exploiting this dependency earlier, that is, by replacing the hybrid CNN-transformer encoder with a transformer-only encoder? The vision transformer (ViT) \cite{dosovitskiy2020image}, is competitive against the most performant CNNs in various computer vision tasks. However, it remains largely unexploited in STR \cite{atienza2021vision}.

%most of the state-of-the-art methods in STR have been utilizing a hybrid CNN-transformer as the encoder \cite{fang2021read, wang2021two} and using ViT as a backbone in STR models is largely unexploited \cite{atienza2021vision}.

We discovered that employing ViT as an encoder followed by a transformer decoder gives competitive result in STR. However, there are two areas to improve on. First, ViT uses a linear layer to project image patches into encodings. The analysis in \cref{sec:areas} shows that different patch resolutions can have detrimental impact on scene text images of certain word lengths and resizing scales. This finding may apply to other architectures that utilize patches. 

Second, transformer decoder employs an autoregressive decoding process and therefore, lesser information is available to leading decoded characters as compared with trailing ones. Our analysis indicates that the first character, which is decoded without any information from previous character, has the highest error rate. This may also be prevalent in other autoregressive methods.
%, among all the characters if conditioned on ground truth previous characters. This error rate is observed to decline over trailing characters

To address the aforementioned areas, we propose a transformer-only model that can process different patch resolutions and decode in both the original and reverse character orders (e.g ‘boy’ and ‘yob’). Inspired by the mixture of experts, we call this technique integrated experts. The model can effectively represent scene text images of multiple resizing scales. It also complements autoregressive decoding with minimal additional latency as opposed to ensemble. 

% \jj{To address the two issues, we propose to add the following features to the baseline model of a ViT encoder and a transformer decoder: 1) The encoder can accept image patches of any size (not of only one size like the original ViT), 2) the decoder can produce the target output in any direction (either left-to-right or right-to-left), and 3) the model is trained with all combinations of the patch size and the decoding direction, as multi-task learning. The multi-task learning helps the model be trained to learn dependencies from different patch sizes and from both decoding directions.}\jjc{is it accurate?}

In summary, the contribution of this work is as follows: (1) a strong transformer-only baseline model, (2) identification of areas for improvement in transformer for STR, (3) the integrated experts method which serves to address the areas for improvement, and (4) state-of-the-art results for $6$ out of the $7$ benchmarks.

The rest of the paper is organized as follows. Firstly, \cref{sec:related_work} explores related works. Secondly, \cref{sec:areas} analyses the areas for improvement in using transformer in STR. Thirdly, \cref{sec:methodology} discusses the proposed methodology. Following which, \cref{sec:experiment} reports the experimental results on $7$ scene text benchmarks. Lastly, \cref{sec:conclusion} concludes this study.

%------------------------------------------------------------------------
\section{Related Work}
\label{sec:related_work}

The encoder-decoder framework is a popular approach in the field of STR \cite{qiao2020seed}. Traditionally, CNN was used to encode scene text images and RNN was used to model sequence dependency and translate the encoded features into a sequence of characters. Shi et al. \cite{shi2016end} proposed a CNN encoder followed by deep bi-directional long-short term memory \cite{hochreiter1997long} for decoding. In a similar work \cite{shi2018aster}, a rectification network was introduced into the encoder in order to rectify the image before features are extracted by a CNN.

As transformer became a de facto standard for sequence modeling tasks, works that incorporate transformer as the decoder are becoming more common in STR. Lu et al. \cite{lu2021master} proposed a multi-aspect global context attention module, a variant of global context block \cite{cao2019gcnet}, as part of the encoder network. A transformer decoder is then used to decode the image features into sequences of characters. A similar model was also proposed by Wu et al. \cite{wu2021naster}, utilizing a transformer decoder which is preceded by a global context ResNet (GCNet). Zhang et al. \cite{zhang2021scene} employed a combination of CNN and RNN as the encoder and a transformer inspired cross-network attention as a part of the decoder in their cascade attention network. Similarly, Yu et al. \cite{yu2020towards} introduced a global semantic reasoning module made up of transformer units, as a module in the decoder. %Using a feature pyramid network (FPN) to aggregate features from a CNN, the visual features are fed into the reasoning module to capture semantic information. 

% Alternatively, Bartz et al. \cite{bartz2019kiss} employed a localization network which predicts region of interest of scene text images. Cropped images from the localization are then passed onto a hybrid CNN-transformer encoder followed by a transformer decoder. 

Apart from being used as/in the decoder, transformer has also been employed in the encoder in the form of a hybrid CNN-transformer \cite{bartz2019kiss}. Fu et al. \cite{fu2021look} proposed the use of hybrid CNN-transformer to extract visual features from scene text images. It is then followed by a contextual attention module, which is made up of a variant of transformer, as part of the decoding process. Lee et al. \cite{lee2020recognizing} likewise utilized a hybrid CNN-transformer encoder and a transformer decoder as their recognition model. In addition, the authors proposed an adaptive $2$D positional encoding as well as a locality-aware feed-forward module in the transformer encoder. With a focus on the positional encoding of transformer, Raisi et al. \cite{raisi20212lspe} applied a $2$D learnable sinusoidal positional encoding which enables the CNN-transformer encoder to focus more on spatial dependencies.

Non-autoregressive forms of transformer decoder were also proposed in various works, coupled with an iterative decoding.  Qiao et al. \cite{qiao2021pimnet} proposed a parallel and iterative decoding strategy on a transformer-based decoder preceded by a feature pyramid network as an encoder. In a similar fashion, Fang et al. \cite{fang2021read} utilized a hybrid CNN-transformer based vision model followed by a transformer decoder with iterative correction. %The vision model will first predict a sequence of characters based solely on an input image. The predictions will then enter the decoder which outputs new predictions in a non-autoregressive manner. Following which, the predictions will be fed to a fusion module to generate a new sequence as part of an iterative process.

As ViT is becoming a more common approach at vision tasks, Ateinza \cite{atienza2021vision} proposed ViT as both the encoder and non-autoregressive decoder to streamline the encoder-decoder framework of STR. The ViT is made up of the transformer encoder, where the word embedding layer is replaced with a linear layer. By utilizing this one stage process, the author is able to achieve a balance on the accuracy, speed, and efficiency for STR. However, its recognition accuracy does not achieve state-of-the-art performance.%Similarly, Wang et al. \cite{wang2021two} employed a unified structure in place of the two stages encoder-decoder framework. Using transformer as a building block, the authors proposed a masked language-aware module which serves to inculcate linguistic information into their CNN encoder through a weakly-supervised complementary learning. 

%Most related works have adopted transformer as part of their framework with a hybrid CNN-transformer as the encoder, or a transformer as a decoder. Employing ViT as the encoder in the encoder-decoder framework is unexploited and model consisting only of transformer units is largely unexplored. This work utilizes ViT as an encoder with a simple transformer-only, encoder-decoder framework and outperforms the previous works.

%------------------------------------------------------------------------
\section{Areas for Improvement in Transformer}
\label{sec:areas}
\subsection{Encoder: Impact of patch resolution}
\label{sec:encoder}
STR takes cropped images of text from natural scenes as inputs. Therefore, they come in different sizes and aspect ratios. As the images are needed to be of a fixed height and width before being passed as inputs into an STR model, one common approach is to ignore the original aspect ratios and resize them with varying scales. Preserving the original resolutions with padding results in worse performance in the work by Shi et al. \cite{shi2018aster} which is in line with our experimental result (in supplementary material). For ViT, resized images are split into patches, which will be flatten and passed through a linear layer followed by the transformer encoder.

Using a baseline architecture of ViT encoder with transformer decoder as described in \cref{sec:methodology}, several models were trained with different patch resolutions. The distributions of correct predictions were analysed using the relative frequency distribution change \cite{huang2020evolving} as defined in \cref{eq:rel_freq_shift}:
\begin{equation}
  F_{l, s} = \frac{\frac{F^{2}_{l, s} - F^{1}_{l, s}}{F^{1}_{l, s}}}{\frac{\sum_{l, s}(F^{2}_{l, s}-F^{1}_{l, s})}{\sum_{l, s}F^{1}_{l, s}}}
  \label{eq:rel_freq_shift}
\end{equation}
where the subscript $l$ and $s$ represent the word length and scaling factor. The scaling factor defined as $\frac{final\:width}{final\:height}\frac{initial\:height}{inital\:width}$, is the scaling of the initial aspect ratio to the final resized aspect ratio. $F^{1}_{l, s}$ and $F^{2}_{l, s}$ represent the frequency of the correct predictions at word length $l$ with scale factor $s$ of two models. The training dataset specified in \cref{sec:train_dataset} is used to compute $F_{l, s}$ because a large dataset is needed to reliably estimate $F_{l, s}$ at each $l$ and $s$; the number of samples in the benchmark datasets is insufficient.
\begin{figure}
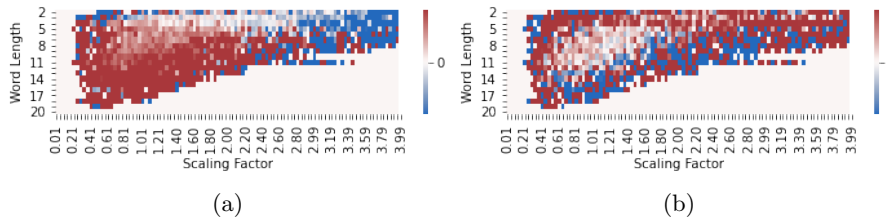

  \centering
    \subfloat[\label{fig:dist-b}]{\includegraphics[width=6cm]{figures/dist_shift_4by8to8by4.pdf}}
    \subfloat[\label{fig:dist-c}]{\includegraphics[width=6cm]{figures/dist_shift_8by4to8by4.pdf}}
  \caption{Relative frequency distribution change in correct predictions (a) from a model trained using patch resolution $4\times8$ to a model trained using $8\times4$. (b) two models trained using input patch resolution of $8\times4$. All models were separately initialized and trained using the same hyperparameters}
  \label{fig:dist}
\end{figure}

\cref{fig:dist} visualizes the relative frequency distribution change, where the word length is ranged from $2$ to $20$ with scaling factor ranging from $0$ to $4$. Bins with frequency count lesser than $100$ are removed. Noting that the remaining count account for $95\%$ of total count, these arrangements will reduce the noise caused by bins with low frequency and provide better visuals. In \cref{fig:dist-b}, $F^{1}_{l, s}$ and $F^{2}_{l, s}$ are calculated from the models trained with patch resolution of $4\times8$ and $8\times4$ respectively. In \cref{fig:dist-c}, $F^{1}_{l, s}$ and $F^{2}_{l, s}$ are computed with two randomly initialized models trained with the same patch resolution of $8\times4$. As the denominator in \cref{eq:rel_freq_shift} for \cref{fig:dist-b} and \cref{fig:dist-c} is positive, $F_{l, s} > 0$ signifies that $F^{2}_{l, s}$ produces more correct predictions at $l$ and $s$ than $F^{1}_{l, s}$ and vice-versa. 

As plotted in \cref{fig:dist-b}, the two models show clear contrast in terms of performance with respect to word length and scaling factor. In specifics, images with word length 3-5 and scaling factor of 1.2-2.4 are least affected by the patch resolution used (white region in \cref{fig:dist}. Images with (1) word length of 2-3, scaling factor $<$ 1; and (2) word length 2-11, scaling factor $>$ 2.6, favours patch resolution of $4\times8$ (blue regions). The red region represents images that performs better with $8\times4$. These findings suggest that models trained with different resolutions are experts for certain word lengths and scales. Furthermore, \cref{fig:dist-c} shows no distinct contrast in the frequency between the two separately initialized models (trained with same patch resolution) as opposed to \cref{fig:dist-b}. This provides a stronger evidence for the impact of different patch resolutions in STR.

%Similarly, two models of the same hyperparameters were trained using patch sizes of $8\times4$ and $4\times8$. The relative frequency shift of correct predictions for $8\times4$ from $4\times8$ is illustrated in\cref{fig:dist-c}. As observed, the patch size of $8\times4$ performs relatively better than $4\times8$ on images that require a smaller scaling factor for the resizing. Meanwhile, performance of the model on words with shorter length and/or larger scaling factor tend to be better with a $4\times8$ patch. These are in contrast to \cref{fig:dist-b} where no distinct pattern is seen. It seems that a higher width resolution for the patch size is more suited for images that require larger scaling. Conversely, a higher height resolution is better for images that are required to be scaled down. The finding suggests that different resolutions with the same patch size for the model fills a different `niche' in STR.

\subsection{Decoder: Errors in first character prediction}
\label{sec:decoder}

\begin{figure}[htbp]
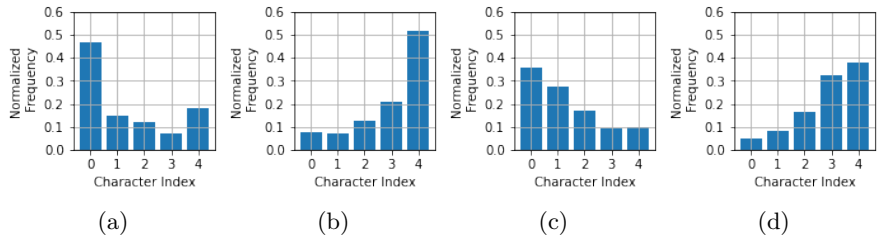

  \centering
  \subfloat[\label{fig:freq-a}]{\includegraphics[width=0.24\linewidth]{figures/freq_one_wrong_LR.pdf}}
  \subfloat[\label{fig:freq-b}]{\includegraphics[width=0.24\linewidth]{figures/freq_one_wrong_RL.pdf}}
  \subfloat[\label{fig:freq-c}]{\includegraphics[width=0.24\linewidth]{figures/freq_two_wrong_LR.pdf}}
  \subfloat[\label{fig:freq-d}]{\includegraphics[width=0.24\linewidth]{figures/freq_two_wrong_RL.pdf}}
  \caption{Normalized frequency distributions of wrong predictions for word length $5$ at the character indices, conditioned on ground truth characters. (a) Predictions with one wrong character. (b) Predictions with one wrong character trained on reversed ground-truths. (c) Predictions with two wrong characters. (d) Predictions with two wrong characters trained on reversed ground-truths}

  \label{fig:char_wrong_pred}
\end{figure}

Two baseline models as described in \cref{sec:approach} were randomly initialized and trained separately where one of them uses the original ground-truth texts while the other uses reversed ground-truths. Our experimental results for wrong predictions on train dataset are plotted in \cref{fig:char_wrong_pred}. It is to be noted that the incorrect predictions used to plot \cref{fig:char_wrong_pred} are words with length $5$ where there is only one incorrectly predicted character for \cref{fig:freq-a} and \cref{fig:freq-b}, and two incorrectly predicted characters for \cref{fig:freq-c} and \cref{fig:freq-d}. 

In \cref{fig:freq-a} and \cref{fig:freq-c}, the first decoded character is at index $0$. Whereas in \cref{fig:freq-b} and \cref{fig:freq-d}, the order of character indices was flipped to reflect the reversed ground-truth texts. In the latter case, index $4$ would be the first decoded character. As the transformer decoder is autoregressive, the predictions are conditioned on ground truth characters in order to evaluate the accuracy on individual character given the correct prior character(s).

The experimental results show that both models have the highest error rate when decoding the first character, and such observations can be seen in other word lengths as well as other numbers of incorrect characters. Also, characters that are decoded subsequently tend to have lower error rates, given the correct previous characters inputs. More analysis is in the supplementary material.

%We hypothesize that these may be due to less information being available when decoding earlier characters, as compared with later characters, in the autoregressive decoding process.

%------------------------------------------------------------------------

\begin{figure}[!ht]
  \centering
    \includegraphics[width=1\textwidth]{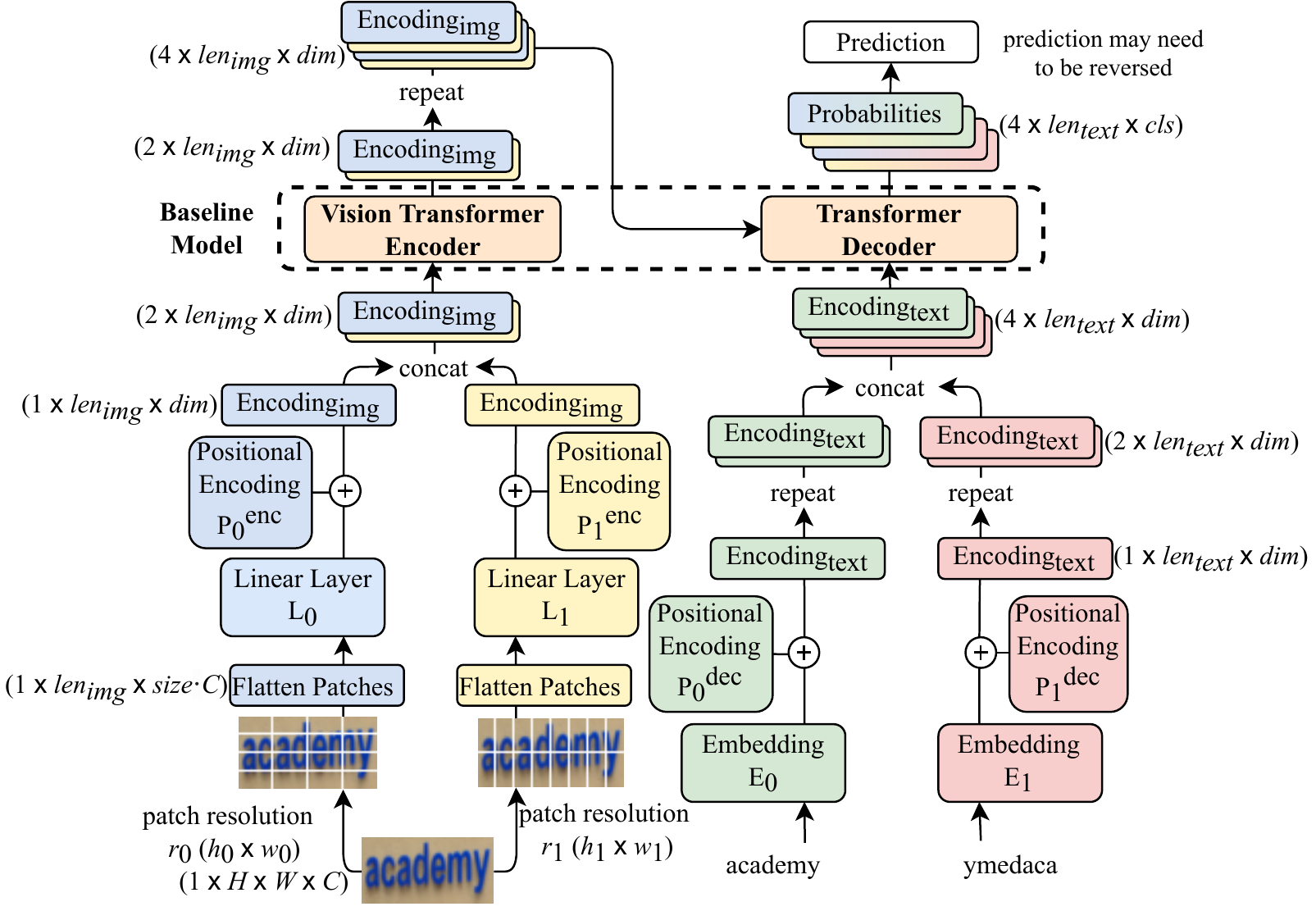}
  \caption{Architecture of PTIE. It is to be noted that there is no attention between the concatenated Encoding\textsubscript{img}. This is also the case for concatenated Encoding\textsubscript{text}. The attention is utilized as per vanilla transformer}
  \label{fig:architecture}
\end{figure}

\section{Methodology}
\label{sec:methodology}

\subsection{Model Architecture and Approach}
\label{sec:approach}

Architecture of the proposed baseline model is illustrated in \cref{fig:architecture}. It consists of a ViT encoder and a transformer decoder. Inspired by the mixture of experts, we present a transformer with integrated experts, named PTIE, to improve on the areas discussed in \cref{sec:areas}. Each expert, denoted as $Exp_{i, j}$, requires image patches of resolution $r_i$ and ground-truth texts of type $j$ to be trained where $i,j \in \{0,1\}$. For this work, patch resolution $r_0$ has the dimension of $h_0\times w_0=4\times8$, and $r_1$ has that of $h_1\times w_1=8\times4$. Both patch resolutions have the same patch size $size=h_0w_0=h_1w_1$. $j=0$ represents the use of original ground-truth texts (e.g. `academy'), and $j=1$ for reversed ground-truths (e.g. `ymedaca').

For expert $Exp_{i, j}$, the resized image of dimension $H\times W\times C$ (height $\times$ width $\times$ channels) will first be split up into patches of resolution $r_i$ and then flatten. The sequence of flatten patches with length of $len_{img}$ is passed through linear layer $L_i$. The output, Encoding\textsubscript{img}, will then be summed with positional encoding $P^{enc}_i$ before going through the encoder with encoding dimension of $dim$. Similarly, the ground-truths of type $j$ will go through an embedding layer $E_j$ and outputs Encoding\textsubscript{text} with sequence length, $len_{text}$. It is then summed with $P^{dec}_j$ before being passed into the decoder which produces the probabilities over the total number of classes, $cls$. Cross-entropy loss will then be applied to the probabilities with their respective type $j$ ground-truths.

In our design, all experts are integrated into 1 model. The parameters in the encoder and decoder are shared. The differentiating factors among them are the initial linear/embedding layers as well as the positional encodings. More precisely, each expert shares about $96\%$ of the parameters with the others and each sample from the dataset will have 4 sets of input (1 for each expert), namely: (1) image split into patches of $4\times8$ with the original ground-truth text, (2) $4\times8$ patches with reversed ground-truth text, (3) $8\times4$ patches with original ground-truth, and (4) $8\times4$ patches with reversed ground-truth. It is to be noted that our baseline model mentioned in this work employs only 1 set of input (e.g. $i=0,j=0$: $4\times8$ patches with original ground-truth).

The manipulation of the dimensions with repeat and concatenation depicted in \cref{fig:architecture} ensures that PTIE decodes each sample image only once despite having 4 sets of initial input. This will allow the inference latency to be close to that of the baseline model. As an ensemble-inspired method, the model will generate 4 predictions for a given sample. The output with the highest word probability (calculated by the multiplications of characters probability) will then be selected as the final prediction. However, different from a standard ensemble, our proposed model requires only a quarter of the parameters and inference time while remaining competitive against an ensemble of models in terms of accuracy.

\subsection{Positional Encoding}
\label{sec:PE}
According to the study by Ke et al \cite{ke2020rethinking}, the positional encoding used in the vanilla transformer \cite{vaswani2017attention} causes noisy correlation with the embeddings of input tokens (e.g. characters) and may be detrimental to the model. Therefore, on top of the aforementioned proposed model, their strategy of untying the positional encoding from the input token embedding was also adopted.

Instead of summing the positional encoding, a positional attention is instead calculated and then added during the multi-head attention process. The positional attention for the encoder, $\alpha^{enc}_i$, is calculated as in \cref{eqn:enc_pos}:
\begin{equation}
    \label{eqn:enc_pos}
    \alpha^{enc}_i= \frac{1}{\sqrt{2d}}(P^{enc}_{i}W_Q)(P^{enc}_{i}W_K)^T
\end{equation}
where $P^{enc}_{i}$ is the positional encoding of the patches with resolution $r_i$; $d$ is the dimension of positional encodings; $W_Q$ and $W_K$ are linear layers with the same number of input and output dimensions. All layers in the encoder share the $\alpha^{enc}_i$.

The decoder has masked self-attention and cross-attention layers. Their positional attentions, $\alpha^{dec}_j$ and $\alpha^{dec\_c}_{i,j}$, are calculated as in \cref{eqn:dec_m_pos} and \cref{eqn:dec_pos} respectively:
\begin{equation}
    \label{eqn:dec_m_pos}
    \alpha^{dec}_j= \frac{1}{\sqrt{2d}}(P^{dec}_{j}U_Q)(P^{dec}_{j}U_K)^T
\end{equation}

\begin{equation}
    \label{eqn:dec_pos}
    \alpha^{dec\_c}_{i,j}= \frac{1}{\sqrt{2d}}(P^{dec}_{j}V_Q)(P^{enc}_{i}V_K)^T
\end{equation}
where $i$ and $j$ denote the types of patch resolution and ground-truth and $U_Q$, $U_K$, $V_Q$, $V_K$ are linear layers like $W_Q$ and $W_K$. Similarly, all the layers in the decoder share the same positional attentions.

\begin{figure}[htbp!]
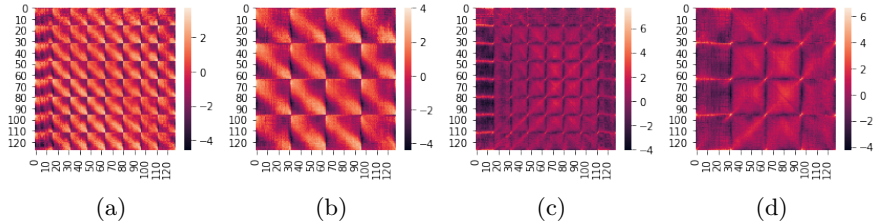

  \centering
  \subfloat[\label{fig:pos_att_4by8_10}]{\includegraphics[width=0.24\linewidth]{figures/pos_att_4by8_10.pdf}}
  \subfloat[\label{fig:pos_att_4by8_16}]{\includegraphics[width=0.24\linewidth]{figures/pos_att_8by4_10.pdf}}
  \subfloat[\label{fig:pos_att_8by4_10}]{\includegraphics[width=0.24\linewidth]{figures/pos_att_4by8_16.pdf}}
  \subfloat[\label{fig:pos_att_8by4_16}]{\includegraphics[width=0.24\linewidth]{figures/pos_att_8by4_16.pdf}}
  \caption{Learned unnormalized positional attention maps in the encoder of PTIE for (a) head 1, resolution=$4\times8$; (b) head 1, resolution=$8\times4$; (c) head 2, resolution=$4\times8$; (d) head 2, resolution=$8\times4$. The axes represent the indices of the flatten patches}
  \label{fig:pos_att}
\end{figure}

The image patches of both resolutions were flatten in row-major order. With a large amount of parameters sharing, the spatial layouts of flatten patches with different patch resolutions for PTIE are handled by the positional encodings as shown in \cref{fig:pos_att}. Thus, the unnormalized positional attention maps for patch resolution $4\times8$ and $8\times4$ are different.

\subsection{Implementation Details}
\label{ImplementationDetail}
The network was implemented using PyTorch and trained with ADAM optimizer with a base learning rate of $0.02$, betas of $(0.9, 0.98)$, and eps of $1e^{-9}$, warmup duration of $6000$ steps with a decaying rate of $\min({steps^{-0.5}}, steps \times  warmup^{-1.5})$. The models were trained on $5$ NVIDIA RTX3090, with a batch size of $640$. All experiments were trained for $10$ epochs. Images are grayscaled and resized to a height and width of $32$ by $128$ without retaining the original aspect ratios. Standard augmentation techniques following Fang et al.'s work \cite{fang2021read} were applied. The models in all experiments contain $6$ encoder layers and $6$ decoder layers with dropout of $0.1$. The encoding dimension is $512$ with $16$ heads for the multi-head attention. The feed forward layer has an intermediate dimension of $2048$. The model recognizes $100$ classes for training, including $10$ digits, $52$ case sensitive alphabets, $35$ punctuation characters, a start token, an end token, and a pad token. For testing, only $36$ case-insensitive alphanumeric characters were taken into consideration as per related works \cite{shi2018aster, yan2021primitive, fang2021read}. Greedy decoding was used with a maximum sequence length of $30$. No rotation strategy \cite{li2019show} was used.

%------------------------------------------------------------------------
\section{Experimental Results and Analysis}
\label{sec:experiment}

\subsection{Datasets}
\subsubsection{Synthetic datasets.}
\label{sec:train_dataset}
Two synthetic datasets were used: \emph{MJSynth} (MJ) \cite{jaderberg2014synthetic}, with $9$ million samples, and \emph{SynthText} (ST) \cite{Gupta16}, containing 8 million images. Some works utilized \emph{SynthAdd} (SA) \cite{li2019show} due to the lack of punctuation in \emph{MJSynth} and \emph{SynthText}. SA was not used in our training.

\subsubsection{Real datasets.}
For evaluation, $6$ datasets of real scene text images which contain $7$ benchmarks were used. \emph{IIIT 5K-Words} (IIIT5K) \cite{mishra2012scene} contains $3000$ test images. \emph{ICDAR 2013} (IC13) \cite{karatzas2013icdar} contains $1015$ testing images as per related work \cite{wang2011end}. Two verions of \emph{ICDAR 2015} (IC15) \cite{karatzas2015icdar} , containing $2077$ test images and $1811$ images, were used for evaluation. \emph{Street View Text} (SVT) \cite{wang2011end} consists of $647$ testing images. \emph{Street View Text-Perspective} (SVT-P) \cite{phan2013recognizing} contains $645$ testing images. \emph{CUTE80} (CT) \cite{risnumawan2014robust} contains $288$ test images. \emph{COCO-Text} \cite{gomez2017icdar2017} which contains $42,618$ training images were used for fine-tuning so as to compare with works which uses real datasets in training or fine-tuning.

%Among the benchmark datasets above, IC15, SVT-P, and CUTE80 are classified as irregular text datasets as most of the images contain certain form of distortions. IIIT5K, IC13, and SVT are classified as regular text datasets.

\subsection{Comparison with State-of-the-Art Methods}

\begin{table}[ht!]
\scriptsize
\begin{center}
\caption{Comparison of accuracies on benchmark datasets with works trained using synthetic datasets. PTIE--Untied uses the learnable positional encoding discussed in \cref{sec:PE} while PTIE--Vanilla uses it as per vanilla transformer method. The best and second best results as compared with PTIE--Untied are in bold and underline respectively. Values in the parenthesis are the difference in accuracy between the proposed model with the best or next best result. Note that the comparison of results are only between a PTIE-based model and other related works}
\label{tab:result}

\begin{tabular}{|c|c|c|c|c|c|c|c|c|c|}
\hline
{\bf{Method}} & \bf{Year} & \bf{Train} & \multicolumn{3}{c|}{\bf{Regular Text}} & \multicolumn{4}{c|}{\bf{Irregular Text}} \\ \cline{4-10}
& & \bf{Datasets} & IIIT & IC13 & SVT & \multicolumn{2}{c|}{IC15} & SVT-P & CT \\ \cline{4-10}
& & & 3000 & 1015 & 647 & 2077 & 1811 & 645 & 288 \\
\hline\hline
Luo et al. \cite{LUO2019109} & PR `19 & ST+MJ & 91.2 & 92.4 & 88.3 & 68.8 & - & 76.1 & 77.4 \\ 
Yang et al. \cite{yang2019symmetry} & ICCV `19 & ST+MJ & 94.4 & 93.9 & 88.9 & 78.7 & - & 80.8 & 87.5 \\
Zhan and Lu \cite{zhan2019esir} &  CVPR `19 & ST+MJ & 93.3 & 91.3 & 90.2 & 76.9 & - & 79.6 & 83.3 \\
% Bartz et al. \cite{bartz2019kiss}& 2019 & ST+MJ+SA & 94.6 & 93.1 & 89.2 & 74.2 & - & 83.1 & 89.6 \\
Wang et al. \cite{wang2020decoupled} & AAAI `20 & ST+MJ & 94.3 & 93.9 & 89.2 & 74.5 & - & 80.0 & 84.4 \\
Wan et al. \cite{wan2020textscanner} & AAAI `20 & ST+MJ & 93.9 & 92.9 & 90.1 & - & 79.4 & 84.3 & 83.3 \\
Zhang et al. \cite{zhang2020autostr} & ECCV `20 & ST+MJ & 94.7 & 94.2 & 90.9 & - & 81.8 & 81.7 & - \\
Yue et al. \cite{yue2020robustscanner} & ECCV `20 & ST+MJ & 95.3 & 94.8 & 88.1 & 77.1 & - & 79.5 & 90.3 \\
Lee et al. \cite{lee2020recognizing} & CVPRW `20 & ST+MJ & 92.8 & 94.1 & 91.3 & 79.0 & - & 86.5 & 87.8 \\
Yu et al. \cite{yu2020towards} & CVPR `20 & ST+MJ & 94.8 & - & 91.5 & - & 82.7 & 85.1 & 87.8 \\
Qiao et al. \cite{qiao2020seed} & CVPR `20 & ST+MJ & 93.8 & 92.8 & 89.6 & 80.0 & - & 81.4 & 83.6 \\
Lu et al. \cite{lu2021master} & PR `21 & ST+MJ+SA & 95.0 & 95.3 & 90.6 & 79.4 & - & 84.5 & 87.5 \\
Raisi et al. \cite{raisi20212lspe} & CRV `21 & ST+MJ & 94.8 & 94.1 & 90.4 & 80.5 & - & 86.8 & 88.2 \\
Qiao et al. \cite{qiao2021pimnet} & ACMMM `21 & ST+MJ & 95.2 & 93.4 & 91.2 & 81.0 & 83.5 & 84.3 & 90.9 \\
Atienza \cite{atienza2021vision} & ICDAR `21 & ST+MJ & 88.4 & 92.4 & 87.7 & 72.6 & 78.5 & 81.8 & 81.3 \\
Zhang et al. \cite{zhang2021spin} & AAAI `21 & ST+MJ & 95.2 & 94.8 & 90.9 & 79.5 & 82.8 & 83.2 & 87.5 \\
Wang et al. \cite{wang2021two} & ICCV `21 & ST+MJ & 95.8 & 95.7 & 91.7 & - & 83.7 & 86.0 & 88.5 \\
Wu et al. \cite{wu2021naster} & ICMR `21 & ST+MJ & 95.1 & 94.4 & 90.7 & - & 84.0 & 85.0 & 86.1 \\
Fu et al. \cite{fu2021look} & ICMR `21 & ST+MJ & \underline{96.2} & \underline{97.3} & 93.5 & - & 84.9 & 88.2 & 91.2 \\
Zhang et al. \cite{zhang2021scene} & ICMR `21 & ST+MJ & 90.3 & 96.8 & 89.5 & 76.0 & - & 78.5 & 78.9 \\ 
Luo et al. \cite{luo2021separating} & IJCV `21 & ST+MJ & 95.6 & 96.0 & 92.9 & \underline{81.4} & 83.9 & 85.1 & \underline{91.3} \\
Yan et al. \cite{yan2021primitive} & CVPR `21 & ST+MJ & 95.6 & - & \underline{94.0} & - & 83.0 & 87.6 & \bf{91.7}\\
Baek et al. \cite{baek2021if} & CVPR `21 & ST+MJ & 92.1 & 93.1 & 88.9 & 74.7 & - & 79.5 & 78.2 \\
Fang et al. \cite{fang2021read} & CVPR `21 & ST+MJ & \underline{96.2} & \bf{97.4} & 93.5 & - & \underline{86.0} & \underline{89.3} & 89.2 \\
\hline
PTIE--Vanilla &  & ST+MJ & 96.7 & 97.1 & 95.5 & 83.4 & 87.4 & 89.8 & 91.3 \\
& &  & \textcolor{ForestGreen}{(+0.5)} & \textcolor{red}{(-0.3)} & \textcolor{ForestGreen}{(+1.5)} & \textcolor{ForestGreen}{(+2.0)} & \textcolor{ForestGreen}{(+1.4)} & \textcolor{ForestGreen}{(+0.5)} & \textcolor{red}{(-0.4)}\\
\hline
PTIE--Untied &  & ST+MJ & \bf{96.3} & 97.2 & \bf{94.9} & \bf{84.3} & \bf{87.8} & \bf{90.1} & \bf{91.7} \\
& &  & \textcolor{ForestGreen}{(+0.1)} & \textcolor{red}{(-0.2)} & \textcolor{ForestGreen}{(+0.9)} & \textcolor{ForestGreen}{(+2.9)} & \textcolor{ForestGreen}{(+1.8)} & \textcolor{ForestGreen}{(+0.8)} & (0.0)\\
\hline

\end{tabular}
\end{center}
\end{table}

\begin{figure}[ht!]
  \centering
    \includegraphics[width=0.9\linewidth]{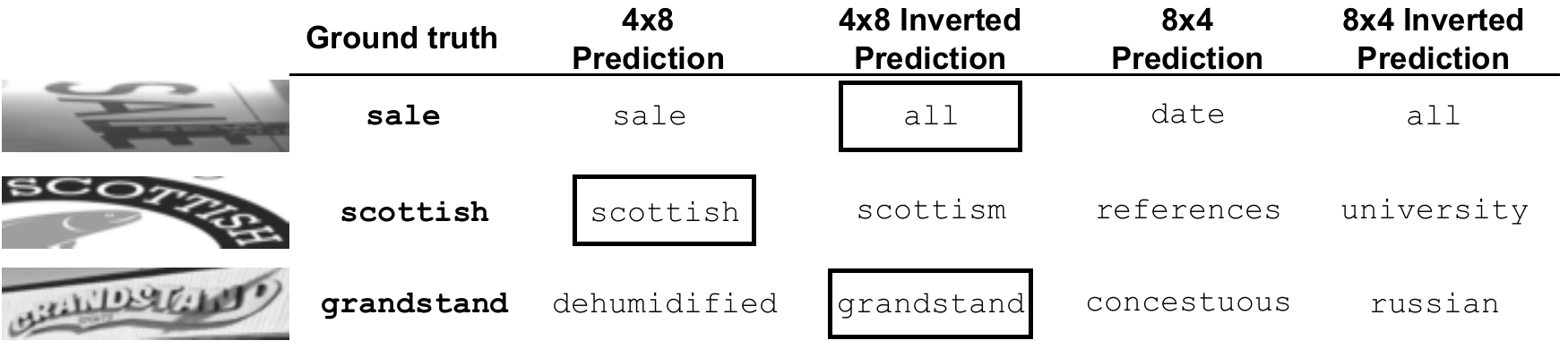}
  \caption{Sample images of success and failure cases. The boxed text represents final output from PTIE. More examples are in the supplementary material}
  \label{fig:cases}
\end{figure}

The results of PTIE are compared with recent works from top conferences and journals as shown in \cref{tab:result}. PTIE achieves state-of-the-art results for most of the benchmarks, even though it has a simple architecture. In particular, PTIE--Untied attained the best results in $6$ out of $7$ benchmarks, outperforming the next best method by $0.9\%$ for SVT, $2.9\%$ for IC15 ($2077$), $1.8\%$ for IC15 ($1811$), and $0.8\%$ for SVTP. The model loses out to the best accuracy \cite{fang2021read} on IC13 by $0.2\%$ and achieved the third highest accuracy. Similarly, PTIE--Vanilla attained the highest accuracy in $5$ benchmarks as compared with recent works. \cref{fig:cases} shows examples of success and failure cases.
% and achieves the best accuracy for CT without margin. Surprisingly, the proposed model with such a simple architecture is able to attain overwhelming results.

Comparing with works that utilize real datasets, we fine-tune our PTIE models with real dataset (COCO-Text \cite{gomez2017icdar2017}) with results shown in \cref{tab:result-real}. Through fine-tuning, our proposed model attained some improvement in performance. The model is able to outperform the state-of-the-art methods for $4$ of the benchmarks and achieved the second highest for $2$ bechmarks. Between the PTIE models trained with ST+MJ, PTIE--Untied has a weighted average (over the benchmarks) of $0.1\%$ higher than PTIE--Vanilla. For ST+MJ+R, PTIE--Untied has a weighted average of $0.4\%$ higher than PTIE--Vanilla.

\begin{table}
\begin{center}
\caption{Comparison of accuracies on the benchmark datasets. The letter `R' denotes the use of real dataset either in training or fine-tuning. The best and second best results in comparison with PTIE--Untied are in bold and underline respectively. Values in the parenthesis are the difference in accuracy between the proposed model with the best or next best result. Note that the comparison of results are only between a PTIE-based model and other related works}
\label{tab:result-real}
\scriptsize
\begin{tabular}{|c|c|c|c|c|c|c|c|c|c|}
\hline
{\bf{Method}} & \bf{Year} & \bf{Train} & \multicolumn{3}{c|}{\bf{Regular Text}} & \multicolumn{4}{c|}{\bf{Irregular Text}} \\ \cline{4-10}
& & \bf{Datasets} & IIIT & IC13 & SVT & \multicolumn{2}{c|}{IC15} & SVT-P & CT \\ \cline{4-10}
& & & 3000 & 1015 & 647 & 2077 & 1811 & 645 & 288 \\
\hline\hline

Li et al. \cite{li2019show} & AAAI `19 & ST+MJ+SA+R & 95.0 & 94.0 & 91.2 & 78.8 & - & 86.4 & 89.6 \\
Yue et al. \cite{yue2020robustscanner} & ECCV `20 & ST+MJ+R & 95.4 & 94.1 & 89.3 & 79.2 & - & 82.9 & \underline{92.4} \\
Wan et al. \cite{wan2020textscanner} & AAAI `20 & ST+MJ+R & 95.7 & 94.9 & 92.7 & - & 83.5 & 84.8 & 91.6 \\
Hu et al. \cite{hu2020gtc} & AAAI `20 & ST+MJ+SA+R & 95.8 & 94.4 & 92.9 & 79.5 & - & 85.7 & 92.2  \\
Qiao et al. \cite{qiao2021pimnet} & ACMMM `21 & ST+MJ+R & \bf{96.7} & 95.4 & \underline{94.7} & \bf{85.9} & \underline{88.7} & \underline{88.2} & \bf{92.7} \\
Baek et al. \cite{baek2021if} & CVPR `21 & R & 93.5 & 92.6 & 87.5 & 76.0 & - & 82.7 & 88.1 \\
Luo et al. \cite{luo2021separating} & IJCV `21 & ST+MJ+R & 96.5& \underline{95.6} & 94.4 & 84.7 & 87.2 & 86.2 & \underline{92.4} \\
\hline
PTIE--Vanilla &  & ST+MJ+R & 96.5 & 96.1 & 96.3 & 84.5 & 89.0 & 91.3 & 88.5 \\
& &  & \textcolor{red}{(-0.2)} & \textcolor{ForestGreen}{(+0.5)} & \textcolor{ForestGreen}{(+1.6)} & \textcolor{red}{(-1.4)} & \textcolor{ForestGreen}{(+0.3)} & \textcolor{ForestGreen}{(+3.1)} & \textcolor{red}{(-4.2)}\\
\hline
PTIE--Untied &  & ST+MJ+R & \underline{96.6} & \bf{96.6} & \bf{95.8} & \underline{85.1} & \bf{89.2} & \bf{92.1} & 91.0 \\
& &  & \textcolor{red}{(-0.1)} & \textcolor{ForestGreen}{(+1.0)} & \textcolor{ForestGreen}{(+1.1)} & \textcolor{red}{(-0.8)} & \textcolor{ForestGreen}{(+0.5)} & \textcolor{ForestGreen}{(+3.9)} & \textcolor{red}{(-1.7)}\\
\hline
\end{tabular}
\end{center}
\end{table}

\subsection{Ablation Studies}

\subsubsection{Transformer-only Encoder.}
In order to demonstrate the effectiveness of utilizing transformer-only encoder, 2 models were trained. We used a ViT encoder with transformer decoder as the baseline model and added a 45-layer ResNet \cite{shi2018aster} on top for the second model. Both models have the same hyperparameters. Comparison with works of similar architecture and method are given in \cref{tab:result-ViT}. 

\begin{table}[ht!]
\begin{center}
\caption{Comparison of accuracies with related works that are heavily based on transformer. The related works contain slight variations in the transformer architecture as discussed in \cref{sec:related_work}. The reported accuracy is the weighted average over the $6$ benchmarks. The total count of $7672$ includes IC15 (2077) while $7406$ uses IC15 (1811). Note that Lee et al. \cite{lee2020recognizing} uses two convolutional layers}
\label{tab:result-ViT}
\scriptsize
\begin{tabular}{|c|c|c|c|c|c|c|c|c|c|}
\hline
\bf{Method} & \bf{Encoder} & \bf{Decoder} & \bf{Parameters} & \multicolumn{2}{c|}{\bf{Accuracy}} \\ \cline{5-6}
 &  &  &  & 7672 & 7406 \\ 
\hline\hline
Raisi et al. \cite{raisi20212lspe} & ResNet based + Trans. & Trans. & - & 89.5 & -\\ 
Lu et al. \cite{lu2021master} & GCNet based & Trans. & - & 89.3 & - \\
Wu et al. \cite{wu2021naster} & GCNet based + Trans. & Trans. & - & - & 90.7\\
Lee et al. \cite{lee2020recognizing} & CNN based + Trans. & Trans. & 55.0M & 88.4 & - \\
%Atienza \cite{atienza2021vision} & Vision Trans. & - & 85.8M & 83.8 & 85.6 \\
\hline
 & ResNet based + Trans. & Trans. & 67.8M & 85.7 & 87.1\\
 & Vision Trans. & Trans. & 45.8M & \bf{90.9} & \bf{92.8}\\
\hline
\end{tabular}
\end{center}
\end{table}

The transformer-only model outperforms the other works that employ a hybrid CNN-transformer encoder. This shows that competitive results can be achieved with just a pure transformer model. Furthermore, our experimental results show that adding a ResNet on top of the transformer encoder has a lower performance as compared with just using a vision transformer. Overall, the results suggest that exploiting the long-term dependency at an earlier stage in an encoder-decoder framework appears to be beneficial for STR.

\subsubsection{Comparison with Standard Ensemble.}

To evaluate the effectiveness of integrated experts, 4 separate models were each trained with one of the following inputs: (1) $8\times4$ patches with original ground-truth, (2) $8\times4$ patches with reversed ground-truth, (3) $4\times8$ patches with original ground-truth, and (4) $4\times8$ patches with reversed ground-truth. The ensemble of these 4 models is named Ensemble--Diverse and the PTIE trained with the 4 inputs is named PTIE--Diverse. The weighted average accuracies of the models over $6$ benchmarks are tabulated in \cref{tab:result-ensem1}. It is to be noted that untied positional encoding was used in all experiments of this section.

\begin{table}[ht!]
\begin{center}
\caption{Weighted average accuracies of mutilple methods on $6$ benchmark datasets (with $2077$ samples from IC15). The naming convention for the methods starts with the patch resolution (e.g. $8\times4$) followed by the type of ground-truth used. ``orig. GT'' stands for the original ground-truth text, and ``rev. GT'' stands for the reversed ground-truth}
\label{tab:result-ensem1}
\scriptsize
\begin{tabular}{|c|c|c|}
\hline
\bf{Method} & \bf{Parameters} & \bf{Acc} \\ \cline{3-3}
 &  & 7672 \\ 
\hline\hline
$8\times4$, orig. GT & 45.8M & 90.9\\
$8\times4$, invt. GT & 45.8M & 90.0\\
$4\times8$, orig. GT & 45.8M & 90.5\\
$4\times8$, invt. GT & 45.8M & 90.1\\
%Ensemble--Dual GT & 91.6M & 91.8 \\
Ensemble--Diverse & 183.2M & \bf{92.4}\\
\hline
$8\times4$, orig. GT (1) & 45.8M & 90.9\\
$8\times4$, orig. GT (2) & 45.8M & 90.7\\
$8\times4$, orig. GT (3) & 45.8M & 90.5\\
$8\times4$, orig. GT (4) & 45.8M & 90.7\\
Ensemble--Identical & 183.2M & 92.1\\
\hline
%PTIE--Dual GT & 45.9M & 91.3 \\
PTIE--Diverse & 45.9M & \bf{92.4}\\
PTIE--Identical & 45.9M & 91.0\\
\hline
\end{tabular}
\end{center}
\end{table}

Undoubtedly, the ensemble of the models brought about a significant performance boost. However, the improvement in accuracy comes at the price of requiring a greater amount of model parameters. Ensemble--Diverse needing $183.2$M parameters, achieved an accuracy of $92.4\%$. In contrast, PTIE--Diverse is able to achieve the same result of $92.4\%$ with only a quarter of the parameters.

The effectiveness of different patch resolutions and ground-truth types are also analyzed with 4 randomly initialized models trained with patch resolution of $8\times4$ and original ground-truth. Their accuracies are shown in \cref{tab:result-ensem1} together with their ensemble (Ensemble--Identical) and PTIE--Identical. Although there is only one type of ground-truth and patch resolution, PTIE--Identical is still trained with separate positional encoding, linear layers, and embedding layers as per \cref{sec:approach}. From the experimental results, accuracy of Ensemble--Identical is lower than that of Ensemble--Diverse by $0.3\%$ which highlights the effectiveness of using different resolutions and ground-truth types. Furthermore, PTIE--Identical suffers a 1.4\% drop in accuracy indicating that different resolutions and ground-truth types are crucial for PTIE on leveraging the experts through different positional encoding, linear layers, and embedding layers.

% \subsubsection{Effectiveness of Untying Positional Encoding}
% \begin{table}[ht!]
% \begin{center}
% \begin{tabular}{|c|c|c|}
% \hline
% \bf{Method} & \bf{Parameters} & \bf{Acc} \\ \cline{3-3}
%  &  & 7672 \\ 
% \hline\hline
% PTIE--Vanilla & 44.3M & 92.3\\
% PTIE--Untied PE & 45.8M & 92.4\\

% \hline
% \end{tabular}
% \end{center}
% \caption{Results of PTIE with and without untying of positional encoding.}
% \label{tab:result-untie}
% \end{table}

% An experiment was conducted with untied positional encoding due to the theoretical advantages highlighted by Ke et al. \cite{ke2020rethinking} as discussed in \cref{sec:PE}. The accuracy of PTIE with the untying of positional encoding is slightly higher as shown in \cref{tab:result-untie}. However, the result is not definitive in demonstrating that this strategy is beneficial for PTIE.
\subsubsection{Comparison of latency.}
\cref{tab:latency} shows a comparison of latency with other recent works that are open source. To tabulate the latency, inference on the test benchmarks was done with an RTX3090 and batch size of 1. Using 4 sets of well-designed inputs mentioned in \cref{sec:approach}, both PTIE--Diverse and Ensemble--Diverse achieved the highest average accuracy. Furthermore, the latency of 52ms by PTIE--Diverse is comparable to the baseline ($8 \times 4$, orig. GT) and is a quarter of Ensemble--Diverse. This is because PTIE--Diverse decodes only once per sample despite having 4 sets of input, while Ensemble--Diverse needs to decode 4 times.
MLT-19 \cite{nayef2019icdar2019} containing 10,000 real images for end-to-end scene recognition averages 11.2 texts instances per image. Using a batch size of 11, the latency of PTIE is about 11ms per cropped scene text image (averaging to 0.12s per full image). Therefore, it may not be a problem for real-time applications. Furthermore, in situations such as applications in forensic science (e.g. parsing images from suspect's hard disk) or assistance to visually impaired, accuracy would be valued over latency.

\begin{table}[ht!]
\begin{center}
\caption{Inference time and weighted average accuracy of recent works. The total count of 7672 uses IC15 (2077) on top of the 5 other datasets mentioned in \cref{sec:train_dataset}. 7406 uses IC15 (1811) and 7248 uses IC15 (1811) and a filtered version of IC13. The variation in total count is due to other works using varied set of benchmarks}
\label{tab:latency}
\scriptsize
\begin{tabular}{|c|c|c|c|c|c|c|}
\hline
\bf{Method} & \bf{Year} & \multicolumn{3}{c|}{\bf{Avg. accuracy}} & \bf{Parameters} & \bf{Time} \\ \cline{3-5}
 &  & 7672 & 7406 & 7248 & \bf{(mil.)} & \bf{(ms)} \\ 
\hline\hline
Wang et al. \cite{wang2020decoupled} & AAAI `20 & 86.9 & - & - & 18.4 & 22 \\
Lu et al. \cite{lu2021master} & PR `21 & 89.3 & - & - & 54.6 & 53 \\
Fang et al. \cite{fang2021read} & CVPR `21 & - & 92.8 & - & 36.7 & 27 \\
Yan et al. \cite{yan2021primitive} & CVPR `21 & - & - & 91.5 & 29.1 & 29 \\
%Wang et al. \cite{wang2021two} & ICCV `21 & - & 91.3 & - & 32.8 & 17\\
\hline
$8 \times 4$, orig. GT & & 90.9 & 92.8 & 92.2 & 45.8 & 50 \\
Ensemble--Diverse & & \bf{92.4} & 93.7 & \bf{93.8} & 183.2 & 202 \\
PTIE--Diverse & & \bf{92.4} & \bf{94.1} & 93.5 & 45.9 & 52 \\ 
%(one output)  & & 90.9 & 92.2 & 92.2 & 45.9 & 59 \\ 
%(Four outputs) & & 92.4 & 94.1 & 93.5 & 45.9 & 55 \\ 
\hline
\end{tabular}
\end{center}
\end{table}

\begin{figure}[htbp!]
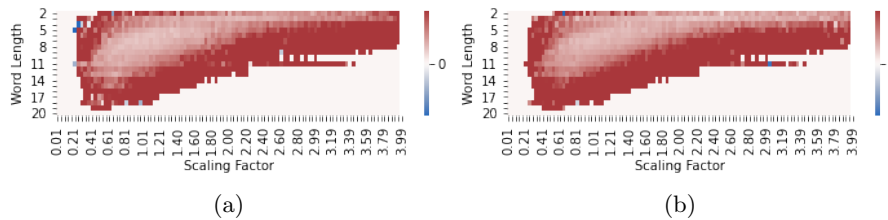

  \centering
    \subfloat[\label{fig:dist_ensem-a}]{\includegraphics[width=6cm]{figures/dist_shift_8by4toEnsem.pdf}}
    \subfloat[\label{fig:dist_ensem-b}]{\includegraphics[width=6cm]{figures/dist_shift_4by8toEnsem.pdf}}
  \caption{Relative frequency distribution change in correct predictions of (a) PTIE from model trained with resolution $4\times8$ and original ground-truth. (b) PTIE from model trained with resolution $8\times4$ and original ground-truth}
  \label{fig:dist_ensem}
\end{figure}

\subsection{Addressing Areas for Improvement}
As per \cref{sec:encoder}, the relative frequency distribution changes of PTIE--Diverse from the models trained with (1) $4\times8$ patches and (2) $8\times4$ patches, are plotted in \cref{fig:dist_ensem-a} and \cref{fig:dist_ensem-b} respectively. Relative improvement in the predictions is seen in most of the lengths and scales for both patch resolutions. This shows that PTIE is effective in utilizing the advantages of both resolutions.

\begin{figure}[htbp!]
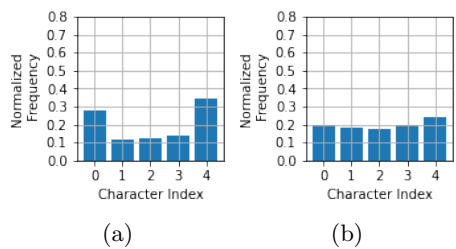

  \centering
  \subfloat[\label{fig:freq_ensem-a}]{\includegraphics[width=0.25\textwidth]{figures/freq_one_wrong_ensem.pdf}}
  \subfloat[\label{fig:freq_ensem-b}]{\includegraphics[width=0.25\textwidth]{figures/freq_two_wrong_ensem.pdf}}
  \caption{Normalized frequency distributions of wrong predictions by PTIE for word length $5$ conditioned on ground truth characters. (a) Predictions with one wrong character. (b) Predictions with two wrong characters}
  \label{fig:char_wrong_pred_ensem}
\end{figure}

Furthermore, the frequency distributions in \cref{fig:char_wrong_pred_ensem} demonstrate that PTIE--Diverse, trained with original and reversed ground-truth, is able to lower the prediction error of first character as discussed in \cref{sec:decoder}. Overall, PTIE is able to improve the accuracy in STR by mitigating the problem of the weak first character prediction. Non-autoregressive decoding is explored in the supplementary material.

%------------------------------------------------------------------------
\section{Conclusion}
\label{sec:conclusion}
In this work, a simple and strong transformer-only baseline was introduced. By exploiting the long-term dependency of STR at an earlier stage in the model, the baseline is able to outperform related works which uses hybrid transformer. We then analyzed and discussed two areas for improvement for transformer in STR. The integrated experts method was proposed to address them and state-of-the-art results were attained for most benchmarks. As the final predictions of PTIE were selected based on word probability, we will explore more selection methods and streamline the processes in PTIE for future work.
\\
\\
\textbf{Acknowledgments:} This work is partially supported by NTU Internal Funding - Accelerating Creativity and Excellence (NTU–ACE2020-03).

\clearpage
% ---- Bibliography ----
%
% BibTeX users should specify bibliography style 'splncs04'.
% References will then be sorted and formatted in the correct style.
%
\bibliographystyle{splncs04}

\end{document}